\pdfoutput=1

\documentclass[11pt]{article}

\usepackage{ACL2023}

\usepackage{times}
\usepackage{latexsym}
\usepackage{multirow}
\usepackage{multicol}
\usepackage[T1]{fontenc}

\usepackage[utf8]{inputenc}

\usepackage{microtype}

\usepackage{inconsolata}

\usepackage{booktabs} 
\usepackage{amsfonts}
\usepackage[T1]{fontenc}
\usepackage{adjustbox}
\usepackage[normalem]{ulem}

\usepackage[utf8]{inputenc}
\usepackage[ruled,vlined]{algorithm2e}
\usepackage{enumitem}
\usepackage{enumerate}
\usepackage{natbib}
\usepackage{cleveref}
\usepackage{subfigure}
\usepackage{booktabs,caption}
\usepackage[flushleft]{threeparttable}
\usepackage{colortbl}
\usepackage{pifont}%
\definecolor{maroon}{cmyk}{0,0.1,0.01,0.01}
\definecolor{blue}{cmyk}{0.1,0.0,0.01,0.01}
\definecolor{yellow}{cmyk}{0.01,0.0,0.2,0.01}

\usepackage{listings}
\usepackage{xcolor}

\definecolor{codegreen}{rgb}{0,0.6,0}
\definecolor{codegray}{rgb}{0.5,0.5,0.5}
\definecolor{codepurple}{rgb}{0.58,0,0.82}
\definecolor{backcolour}{rgb}{0.95,0.95,0.95}

\lstdefinestyle{mystyle}{
    backgroundcolor=\color{backcolour},   
    commentstyle=\color{codegreen},
    keywordstyle=\color{magenta},
    numberstyle=\tiny\color{codegray},
    stringstyle=\color{codepurple},
    basicstyle=\ttfamily\footnotesize,
    breakatwhitespace=false,         
    breaklines=true,                 
    captionpos=t,                    
    keepspaces=true,                 
    numbers=left,                    
    numbersep=5pt,                  
    showspaces=false,                
    showstringspaces=false,
    showtabs=false,                  
    tabsize=2
}

\setlist{nosep}

\newcommand{\ours}{\textsc{PolyIE}\xspace}

\newcommand{\eg}{\emph{e.g.}\xspace} 
\crefname{section}{§}{§§}
\Crefname{section}{§}{§§}

\title{\ours: A Dataset of Information Extraction from Polymer Material Scientific Literature}

\author{Jerry Junyang Cheung$^1$\thanks{These authors contributed equally to this work.} , Yuchen Zhuang$^{1*}$, Yinghao Li$^1$, Pranav Shetty$^2$,\\
\textbf{Wantian Zhao$^1$, Sanjeev Grampurohit$^1$, Rampi Ramprasad$^2$, Chao Zhang$^1$}\\
$^1$College of Computing, $^2$School of Materials Science and Engineering\\
  Georgia Institute of Technology, Atlanta, USA \\
  \texttt{\{jzhang3027,yczhuang,yinghaoli,pranav.shetty,wzhao306,sgrampurohit3\}@gatech.edu} \\
  \texttt{rampi.ramprasad@mse.gatech.edu, chaozhang@gatech.edu}}

\begin{document}
\maketitle
\begin{abstract}
  Scientific information extraction (SciIE), which aims to automatically extract information from scientific literature, is becoming more important than ever.
  However, there are no existing SciIE datasets for polymer materials, which is an important class of materials used ubiquitously in our daily lives.
  To bridge this gap, we introduce \ours, a new SciIE dataset for polymer materials.
  \ours is curated from $146$ full-length polymer scholarly articles, which are annotated with different named entities (i.e., materials, properties, values, conditions) as well as their $N$-ary relations by domain experts.
  \ours presents several unique challenges due to diverse lexical formats of entities,
  ambiguity between entities, and variable-length relations.
  We evaluate state-of-the-art named entity extraction and relation extraction models on \ours,
  analyze their strengths and weaknesses, and highlight some difficult cases for these models.
  To the best of our knowledge, \ours is the first SciIE benchmark for polymer materials, and we hope it will lead to more research efforts from the community on this challenging task.
  Our code and data are available on: \url{https://github.com/jerry3027/PolyIE}.

\end{abstract}

\section{Introduction}

Material science literature is growing at an unprecedented rate.
For example, a simple search on Google Scholar with the term ``polymers'' returns more than 5 million articles on polymer materials.
Such literature reports valuable information on the latest advances in material science, ranging from experimental material properties to material synthesis recipes and procedures.
As machine learning (ML) has achieved success in different applications of material science \cite{butler2018machine, schmidt2019recent}, Scientific Information Extraction (SciIE) from literature for supporting various tasks is becoming increasingly important.
Automatically extracting structured information about materials from massive unstructured literature data can be invaluable to understanding material properties and synthesis, as well as building data-driven ML tools for material discovery~\cite{court2021inverse}.

\begin{figure}[t]
  \centering
  \includegraphics[width=\linewidth]{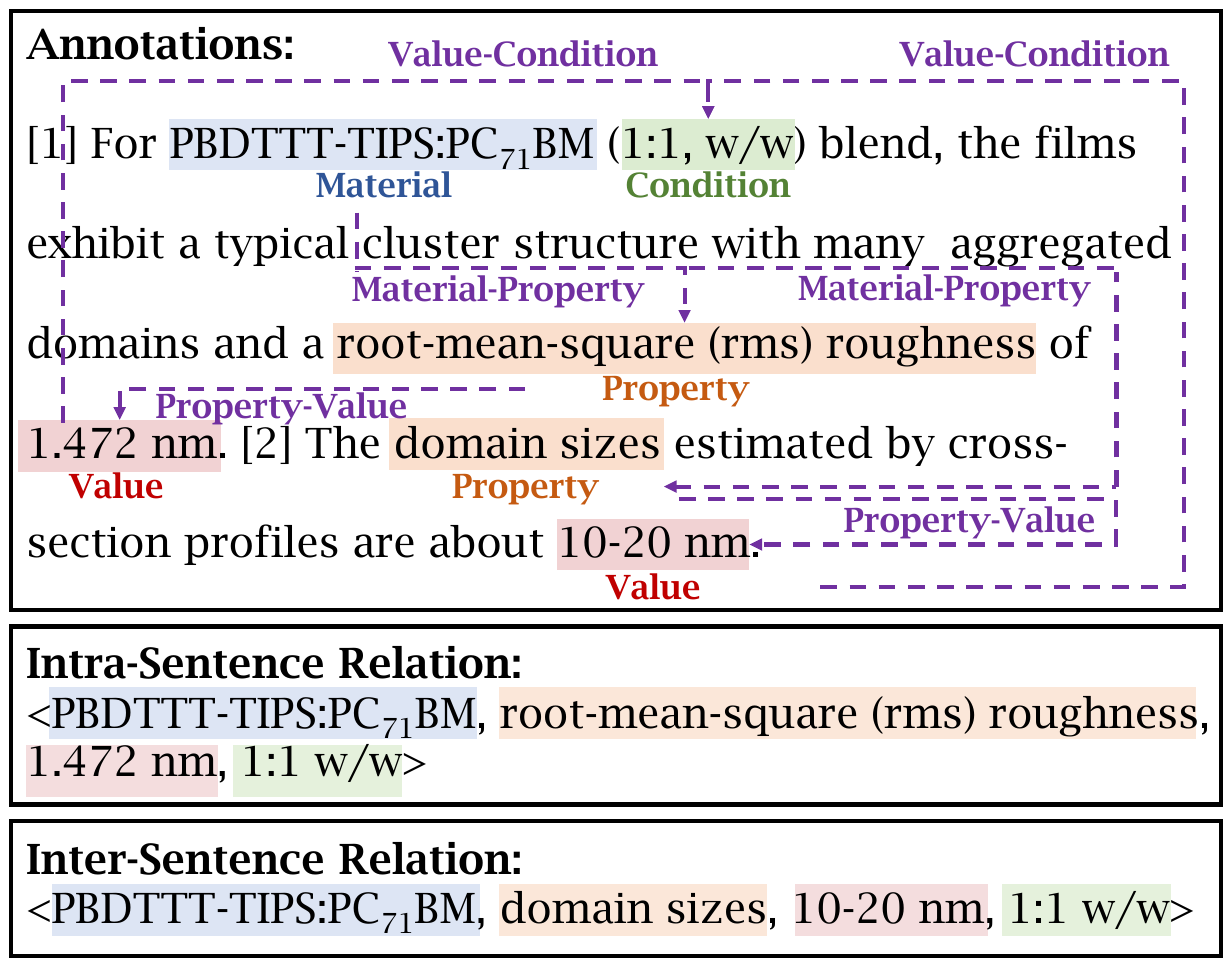}
  \caption{
    An example of entity and relation annotations in \ours from a material science paper~\cite{shi2011copolymer}, including entity mentions as well as intra-sentence and inter-sentence  N-ary relations.}
  \label{fig:intro-example}
\end{figure}

While SciIE has rapidly developed in domains such as biomedical science \cite{luan-etal-2018-multi,gabor2018semeval,jain-etal-2020-scirex,hou-etal-2019-identification,jia-etal-2019-document}, it has made limited progress in the material science domain.
So far, there are only a handful of datasets for material information extraction. 
Some earlier works use ChemDataExtractor~\cite{swain2016chemdataextractor} to automatically generate datasets for battery materials~\cite{huang2020database} and temperatures~\cite{court2018auto}.
More recent datasets are created manually for solid oxide fuel cells~\cite{friedrich-etal-2020-sofc} and material science synthesis procedures~\cite{ogorman-etal-2021-ms}.
However, none of these datasets cover polymer materials, which are an important class of organic materials that play critical and ubiquitous roles in our daily lives.
Due to their versatile properties, polymer materials are being widely used in applications such as packaging, coating, energy saving, and medical applications.
As vast amounts of information on polymer development are being reported in literature data, there is a critical need for SciIE benchmarks and tools to harvest such information from the polymer literature.


To address this gap, we construct a dataset for extracting polymer property knowledge from unstructured literature data.
Our dataset \ours is curated from $146$ full-length polymer scientific articles, which are annotated by domain experts with named entities (i.e., materials, properties, values, conditions) as well as the $N$-ary relations among them (see Figure~\ref{fig:intro-example}). \ours contains $41635$ entity mentions and $4443$ relations in total.
It covers four different application domains of polymer materials: polymer solar cells (PSC), ring-opening polymerization (RP), polymer membranes (PM), and polymers in lithium-ion batteries (LB).
This diversity of content enables the training of models with enhanced generalization capabilities. 
To the best of our knowledge, \ours is the first benchmark for SciIE from full-text polymer literature.

From the natural language processing perspective, extracting information for polymers on \ours introduces unique challenges for both named entity recognition and relation extraction:

\noindent \textbf{Diverse Lexical Formats of Entities.}
Polymer-related entities often have different schemes of nomenclature, such as IUPAC names (\eg, `poly(3-hexylthiophene)'), abbreviations (`PDPPNBr'), trade names (`Styron'), common names (`ABS plastic'), and sample labels (`PE-HDPE-01').
In addition, the identification of polymers can also be achieved through the concatenation of homopolymer names with hyphens or slashes, and the inclusion of numerical values for the component ratios and molecular weights (`PVC-PS-PC-20/30/50-800000').
This diversity of nomenclature in literature poses a challenge for named entity recognition.

\noindent \textbf{Variable-length and Cross-Sentence $N$-ary relations.} Previous research on relation extraction has focused on either binary relations~\cite{luan-etal-2018-multi, yao-etal-2019-docred} or  $N$-ary relations with a fixed number $N$~\cite{jia-etal-2019-document,jain-etal-2020-scirex,zhuang2022resel}.
In contrast, many relations described in polymer literature are variable-length $N$-ary relations.
This is because 1) the reported properties may be describing one or several materials; and 2) different properties can be measured under specific conditions. Furthermore, the elements in a relation tuple may span multiple sentences as shown in Figure~\ref{fig:intro-example}.




We study six mainstay NER and five $N$-ary RE models on \ours in terms of their overall performance and sample efficiency.
We find that the models based on domain-specific pre-trained models (e.g., MatSciBERT) yield better performance than other baselines. However, all the models struggle with accurately recognizing certain categories of named entities and inferring challenging varied-length $N$-ary relations.
Moreover, our observations indicate that, under few-shot settings, the recently popular large language models (LLMs) demonstrate inferior performance than the other baselines on \ours, highlighting potential limitations in comprehending material science concepts.

Our main contributions are: (1) The first polymer information extraction dataset curated from $146$ full-length articles for polymer named entity recognition and relation extraction. (2) Thorough evaluation of seven mainstream NER and five $N$-ary RE models on our curated dataset. (3) Analysis of the difficult cases and limitations of existing models, which we hope will enable future research on this challenging task from the NLP community.

\section{Related Work}

\paragraph{Material Science NLP Datasets.}
Earlier studies~\cite{court2018auto} leverage tools such as ChemDataExtractor~\cite{swain2016chemdataextractor}, ChemSpot~\cite{rocktaschel2012chemspot}, and ChemicalTagger~\cite{hawizy2011chemicaltagger} to perform NER annotation for dataset curation.
For example, ChemDataExtractor is applied to generate datasets for Curie and Neel magnetic phase transition temperatures~\cite{court2018auto} and magnetocaloric materials~\cite{court2021inverse}.
Besides, people also create expert-annotated datasets~\cite{wang-etal-2021-chemner, weston2019named} for the extraction of non-value named entities (\eg, material and property names) and their relationships.
In recent years, there has been an uptick in efforts to include numerical values in datasets for further extraction, with several studies closely related to \ours:
\citet{friedrich-etal-2020-sofc} annotate a corpus of 45 open-access scholarly articles on solid oxide fuel cells, covering entity types of \texttt{materials}, \texttt{values}, and \texttt{devices}.
\citet{panapitiya-etal-2021-extracting} provide annotations of \texttt{CHEM}, \texttt{VALUE}, and \texttt{UNIT} on a set of  papers on soluble materials.
However, both only provide binary relations between pairs of entities, which is inadequate for describing more complex relations. 


\paragraph{$N$-ary Relation Extraction.} $N$-ary relations are size-$N$ tuples that describe the factual relationship between $N$ entities.
In general domains, the MUC dataset~\cite{chinchor1998overview} describes event participants in news articles.
In the biomedical domain, the BioNLP Event Extraction Shared Task~\cite{kim-etal-2009-overview} and PubMed dataset~\cite{jia-etal-2019-document} aim to extract biomedical events from biomedical text.
In the machine learning domain, SciREX~\cite{viswanathan-etal-2021-citationie, jain-etal-2020-scirex, zhuang2022resel} extracts $N$-ary relations in terms of \texttt{<Task, Dataset, Method, Metric>}.
Different from these works' relations, the $N$-ary relations in \ours can have a varied number of named entities, which is more flexible in describing material knowledge but meanwhile introduces new challenges to RE.
The closest work to \ours is drug-combo~\cite{tiktinsky-etal-2022-dataset}, which extracts variable-length combinations of different drugs.
However, \ours and drug-combo are curated for two different domains, and the relations in \ours include  numerical values.


\section{The \ours \ Dataset}

In this section, we describe the details of the \ours dataset. We first
formulate the two information extraction tasks for polymer material literature
in \cref{subsec:data-task}. We then describe the data preprocessing and
annotation procedures in \cref{subsec:data-data} and \cref{subsec:data-annotation}, and finally present the
statistics and characteristics of the \ours dataset in
\cref{subsec:data-quality}.



\subsection{Task Definition}\label{subsec:data-task}

\ours is curated for studying two key information extraction tasks on polymer literature data: (1) identifying relevant named entities, and (2) composing different entities to form  $N$-ary relations.

\paragraph{Named Entity Recognition.} Named Entity Recognition (NER) is the process of locating and classifying unstructured text phrases into pre-defined entity categories such as compound names, property names, etc.
Given a sentence with $n$ tokens $\mathbf{S}=(w_1,\cdots,w_n)$, a named entity mention is a span of  tokens $\mathbf{e}=(w_i,\cdots,w_j)(0\leq i\leq j\leq n)$ associated with an entity type.
In \ours, we focus on NER for describing polymer material properties and include four important entity types: material name, property name, property value, and condition.
An illustrative example can be found in Figure~\ref{fig:intro-example}.
Based on the BIO schema~\cite{li2012joint}, NER can be formulated as a sequence labeling task of assigning a sequence of labels $\mathbf{y}=(y_1,\cdots,y_n)$, each corresponding to a token in the input sentence.

\paragraph{Variable-Length $N$-ary Relation Extraction.} Variable-length $N$-ary relation extraction (RE) refers to the process of identifying and extracting relationships between multiple entity mentions where the number of entities in the relationship can vary.
Formally, given a list of $k$ context sentences $\mathcal{C}=(S_1,\cdots,S_k)$ in one paragraph, let $\mathcal{E}$ be the set of entities appearing in $\mathcal{C}$ where each entity $e \in \mathcal{E}$ belongs to one of the four entity types described in the NER task.
The relation extraction task aims to extract a set of $m$ relations $\mathcal{R}=(r_1,\cdots,r_m)$ from $\mathcal{C}$.
Each relation $r_i$ is a tuple of entities $r_i=(e_1,\cdots,e_{N_i}), (1\leq i\leq m)$ that describe their \texttt{<material, property, value, condition>} relations.
Here, the number of entities $N_i$ can be variable in $\mathcal{R}$ because: 1) the property value may correspond to several materials instead of one; and 2) the condition entity may be absent.
Figure~\ref{fig:intro-example} illustrates this variable-length $N$-ary RE task.



\subsection{Data Preparation}\label{subsec:data-data}

We curate \ours from $146$ publicly available scientific papers, covering four different material science domains: polymer solar cells, ring-opening polymerization, polymer membranes, and lithium-ion batteries.
These papers are sub-sampled from the corpus of 2.4 million material science articles described in \citet{shetty2022general}. This corpus consists of papers published between 2000 to 2021 and is collected from $7$ different material science publishers. Keyword-based search was used to  locate papers that span multiple application domains within polymers. The resulting dataset consists of $100$ papers describing fullerene-acceptor polymer solar cells, $21$ papers describing ring-opening polymerization, $20$ describing lithium-ion batteries, and $5$ describing polymer membranes. The text of these papers is parsed from the PDF of these papers using sciPDF~\footnote{\url{https://github.com/titipata/scipdf_parser}} (a scientific parser based on GROBID~\cite{GROBID}) into utf-8 format. The incorrectly parsed units and symbols are corrected using regular expressions.

\subsection{Data Annotation}\label{subsec:data-annotation}

The \ours dataset is annotated by two polymer science domain experts as well as three computer science graduate students who are trained by the polymer scientists.
Both the NER and RE annotations are performed using the Doccano~\cite{doccano} platform, which is an open-source text annotation tool that facilitates visual annotation with a Web interface.
Below, we detail the annotation schemes for the NER and RE tasks.

\subsubsection{Annotating Named Entities}

In \ours, we annotate mentions of named entities for four categories: material names, property names, property values, and conditions.
Each mention is a continuous text span that specifies the actual name of an entity or its abbreviation.
This is done by marking the entity mention on the Doccano platform with the corresponding entity type.

\noindent \textbf{Compound Names (Material).}
Compound Name entities include text spans that refer to material objects. Only chemical mentions that could be associated with a chemical structure are annotated as Compound Names. They may be specified by a particular composition formula (\eg, ``4,9-di(2-octyldodecyl) aNDT''), a mention of chemical names (\eg, ``trimethyltin chloride''), or just an abbreviation (\eg, ``PaNDTDTFBT''). General chemical nouns (\eg, ``ionic liquids'') are not considered.

\noindent \textbf{Property Names (Property).}
We annotate the properties of chemical compounds as long as they can be measured qualitatively (\eg, ``toxicity'' and ``crystallinity'') or quantitatively (\eg, ``open-circuit voltage'', ``decomposition temperature'').
Corresponding abbreviations should also be annotated (\eg, ``PCE'', ``HOMO level'').

\noindent \textbf{Property Values (Value).}
We annotate the spans that can indicate the degree of qualitative properties (\eg, ``soluble to water'') or describe numerical values with units for quantitative properties (\eg, ``$9.62\times 10^{-5}\ \Omega^{-1}m^{-1}$'', ``5.14 ppm'').

\noindent \textbf{Conditions.}
In material science papers, the properties of materials can be constrained by quantitative modifiers, and we annotate them as conditions to distinguish them from normal property names and property values (\eg, ``room temperature'', ``frequency range 500 Hz -- 3 MHz``).

\subsubsection{Variable-Length $N$-ary Relations} For RE, we annotate the $N$-ary relations between the named entities to capture their \texttt{<Material, Property, Value, Condition>} relations.

\noindent \textbf{Primary Binary Relations.}
As Doccano and most other existing text annotation tools only support annotations for binary relations, we decompose the $N$-ary relation annotation task into simpler binary relation annotation and later aggregate them into full $N$-ary relations. 
We split an $N$-ary relation into multiple binary relations for annotation:
\texttt{Material-Material} marks the relations between material names that constitute one material system; \texttt{Material-Property} identifies the relation between a material and its reported property name; \texttt{Property-Value} annotates the corresponding property name and value; and \texttt{Value-Condition} marks the property values measured under a specific condition.

\noindent \textbf{Transforming Binary to $N$-ary Relations.}
We then transform all the binary relations with common involved entities to generate  $N$-ary relations in the format of \texttt{<Material, (Material), Property, Value, (Condition)>}.
We abandon all binary relations that cannot be combined with other binary relations, only maintaining the generated $N$-ary relations with $N>2$.


\subsubsection{Inter-Annotator Agreement} 
All documents in \ours are annotated by at least two annotators independently. If annotation conflicts arise across two annotators, a third annotator is then assigned to annotate the corresponding sentences independently. The final annotation is determined by majority voting.

We calculate the inter-annotator agreement in terms of Fleiss' Kappa~\cite{fleiss1971measuring}. The Fleiss’ Kappa for individual entity types is calculated by treating other entity types as negative samples. The results are shown in ~\ref{tab:cross-annotator}. The Fleiss' Kappas for \texttt{Material}, \texttt{Property}, and \texttt{Value} are all in the range of almost perfect agreement, while the corresponding value for \texttt{Condition} lies in the range of substantial agreement. For RE, we consider all annotated relations as subjects and treat categories as binary. The Fleiss’ Kappa for RE is 0.67. 

We also compute the average F1-score similar to ~\citet{friedrich-etal-2020-sofc}. The F1-score is calculated by treating one annotator as the gold standard and the other annotator as predicted. For the NER, spans and entity types have to exactly match. For RE, all entity mentions within the n-ary relation have to exactly match. The averaged F1-score for the NER and RE task is 0.89 and 0.84 respectively. 

\begin{table}[h]
\centering
\small
\begin{tabular}{@{}c|c|c|c|c@{}}
\toprule
Overall & Material & Property & Value & Condition \\ \midrule
 0.86    & 0.88  & 0.82 & 0.88 & 0.71    \\ \bottomrule

\end{tabular}
\caption{
   Fleiss' kappa for all annotators across all mentions and each entity type respectively.
  }\label{tab:cross-annotator}
\end{table}


\subsection{Data Analysis}\label{subsec:data-quality}


Table~\ref{table:stats} shows the key statistics for our corpus.
\ours contains $41635$ entity mentions and $4443$ relations in all $146$ fully annotated polymer material science literature.
We quantitively analyze some key properties of \ours:

\noindent \textbf{Statistics of Entities.}
For all the named entity mentions, the distribution of the four entity types \texttt{Material}, \texttt{Property}, \texttt{Value}, and \texttt{Condition} are $49.54\%$, $31.82\%$, $17.00\%$, and $1.70\%$, respectively. In total, those
$41365$ mentions describe
$10890$ distinct named entities
for polymer materials.

\noindent \textbf{Statistics of $N$-ary Relations.}
Among the $4443$ relations on \ours, $86.38\%$ are $3$-ary; $13.62\%$ are $4$-ary; and $3.20\%$ are $5$-ary.
Meanwhile, $26.65\%$ of the relations are cross-sentence relations, while the rest are intra-sentence relations.

\begin{table}[t]
\small
\centering
\fontsize{8}{10}\selectfont\setlength{\tabcolsep}{0.3em}
\begin{tabular}{@{}lccccc@{}}
\toprule
                       & PSC & RP & LB & PM & All \\ \midrule
documents                 & 100  & 21 & 20 & 5 & 146   \\
sentences               & 9,367 & 3,120 & 3,031 & 555 &  16,073        \\
tokens                  & 288,142 & 91,421 & 90,381 & 15,579   & 485,523  \\
avg. tokens/doc.$^*$    & 3,201.6  & 3,102.4 & 3,227.9 & 3,115.8  & 3,325.5 \\ \midrule
mentions                 & 28,775 & 5,760 & 6,013 & 1,087  & 41,635\\
\quad -- \texttt{Material}     & 13,244 & 3,120 & 3,390 & 740  & 20,494       \\
\quad -- \texttt{Property}     & 9,848 & 1,597 & 1,616 & 187 &  13,248      \\
\quad -- \texttt{Value}     & 5,294 & 792 & 835 & 111   &   7,032    \\
\quad -- \texttt{Condition}    & 364 & 150 & 167 & 21 &  702      \\
entities                & 7,099 & 1,621 & 1,739 & 431  & 10,890   \\
avg. mentions/doc.$^*$      & 287.8 & 274.3 & 300.7 & 217.4  & 285.2\\ \midrule
relations               & 3,084 & 592 & 615 & 152  & 4,443  \\
\quad -- 3-ary          & 2,554 & 503 & 516 & 123  &   3,838      \\
\quad -- 4-ary          & 388 & 89 & 99 & 29    &    605   \\
\quad -- 5-ary          & 142 &  - &   - &     - & 142 \\
avg. relations/doc.$^*$  & 30.8 & 28.2 & 30.8 & 30.4  & 30.4  \\ 

\bottomrule
\end{tabular}
\begin{tablenotes}
      \footnotesize
      \item $^*$Avg. indicates average and doc. refers to document.
    \end{tablenotes}
\caption{\ours corpus  statistics.}
\label{table:stats}
\end{table}

\section{Modeling}

In this section, we describe how we model the named entity recognition and $N$-ary relation extraction tasks on \ours.

\begin{figure*}[t]
  \centering
  \includegraphics[width=0.95\linewidth]{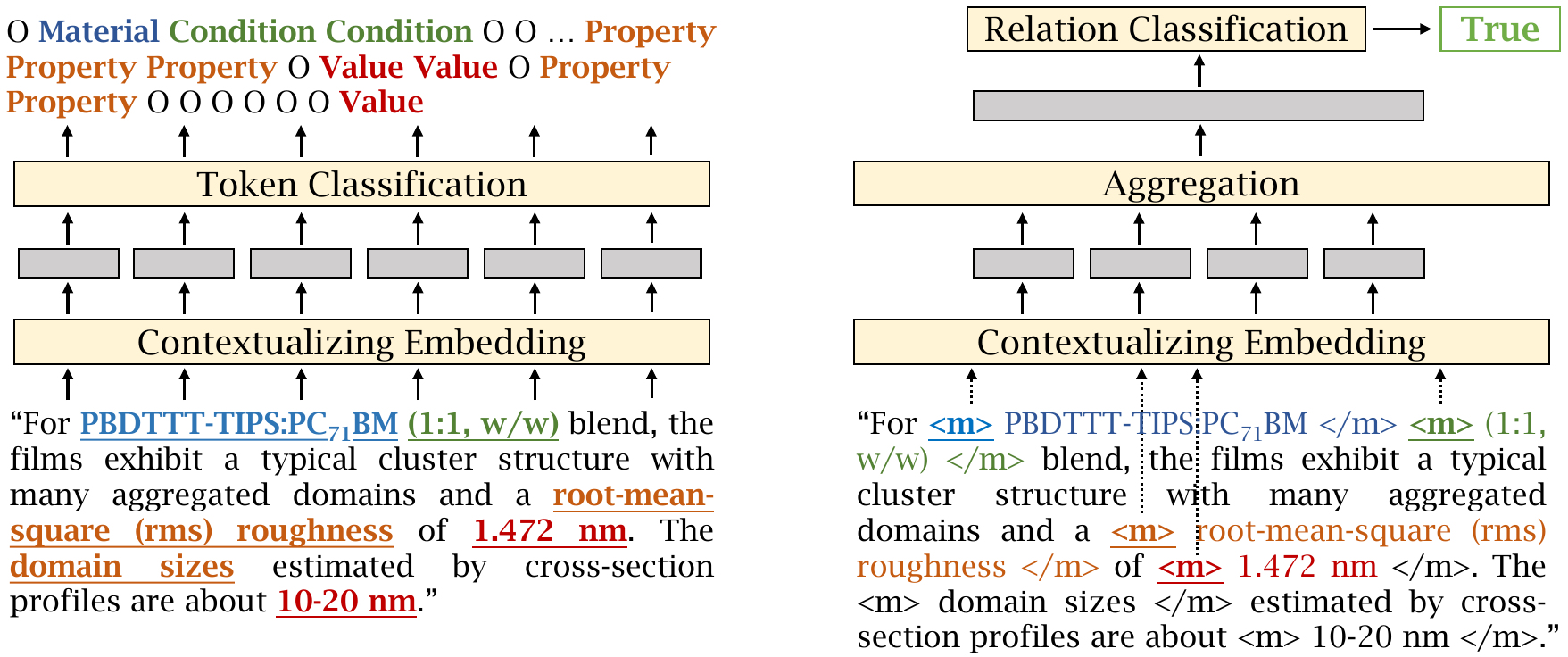}
  \caption{Model architecture for Named Entity Recognition (left) and $N$-ary Relation Extraction (right).}
  \label{fig:model}
\end{figure*}

\paragraph{Named Entity Recognition.} We model the NER task as a sequence labeling problem and learn a neural sequence tagger, as shown in Figure~\ref{fig:model}.
We study both the bi-directional LSTM-CRF (\textbf{BiLSTM-CRF})~\cite{ma-hovy-2016-end} model and BERT-based~\cite{devlin-etal-2019-bert} NER models for neural sequence tagging. We also study the  performance of GPT-3.5 and GPT-4 on NER.

In BiLSTM-CRF, the input text is  passed through an embedding layer to obtain token representations.
These representations are then fed into a BiLSTM layer~\cite{lample-etal-2016-neural} to capture contextual information.
The output of the BiLSTM layer is finally sent to a subsequence Conditional Random Field (CRF) layer~\cite{lafferty2001conditional} for sequence labeling.
For the pre-trained language models (PLM), 
we study both \textbf{BERT$_{\rm base}$} \citep{devlin-etal-2019-bert} and
\textbf{RoBERTa} \citep{liu2019roberta} for NER. 
We also include three domain-specific BERT models: \textbf{SciBERT}~\cite{beltagy-etal-2019-scibert}, \textbf{BioBERT}~\cite{lee2020biobert}, and \textbf{MatSciBERT}~\cite{gupta2022matscibert}. 
All the NER models in the BERT family are fine-tuned for sequence labeling, by stacking a linear layer that maps the contextual token representations into the label space. 
In addition, we also evaluate LLMs' abilities in marking material science concepts. 
Following the existing work~\cite{tang2023does}, We directly prompt \textbf{GPT-3.5-turbo} and \textbf{GPT-4} with few-shot exemplars to use special marks ``@@'' to annotate the boundaries and types of the named entities. Detailed explanations and examples of prompts are included in App.~\ref{app:llm}.

\paragraph{Relation Extraction.} For relation extraction, we evaluate the performances of the rule-based method, PLM-based models, and graph-based models.
For the rule-based method, we leverage the assumption, \textbf{Proximity-Rule}, that relations are more likely to be formed with most proximitive entities.
As illustrated in Figure~\ref{fig:model}, PLM-based models (such as \textbf{BERT-RE}) leverage the strong representation power of pre-trained language models on entities and employ simple aggregation techniques, such as concatenation and summation, to compose relation embeddings for further prediction.
Example models in this category are state-of-the-art models \textbf{PURE}~\cite{zhong-chen-2021-frustratingly},
which inserts a special ``entity marker'' token around the entities in a candidate relation;
and its variant \textbf{PURE-SUM}~\cite{tiktinsky-etal-2022-dataset}, which uses embedding summation for variable-length $N$-ary RE.
We also study graph-based methods for $N$-ary RE, \textbf{DyGIE++}~\cite{luan-etal-2019-general}, which constructs a dynamic span graph from the input text, with entities as nodes and relations as edges to reason over multi-hop relations. 
For models based on LLMs like \textbf{GPT-3.5-turbo} and \textbf{GPT-4}, we randomly choose a subset of examples from the training set to serve as few-shot instances. 
These are then directly sent to the models as prompts to facilitate the relation extraction. 

\section{Experiments}
\begin{table*}[t]\small
  \centering
    \begin{tabular}{l|c|c|c|c|c}
      \toprule
      Model & Material & Property & Value & Condition & Micro Average \\
      \midrule
      BiLSTM-CRF & 58.9 (68.4/51.7) & 70.5 (75.4/66.2) & 73.0 (74.6/71.5) & 13.1 (36.4/8.0) & 65.8 (72.4/60.4)\\\midrule
      BERT$_{\rm base}$ & 83.9 (84.0/83.8) & 77.8 (81.1/74.7) & 81.3 (83.9/79.0) & 13.8 (16.2/12.0) & 80.6 (82.4/78.8) \\
      RoBERTa$_{\rm base}$ & 85.4 (86.4/84.4) & 76.2 (77.4/75.2) & 81.8 (83.3/80.3) & 12.5 (16.7/10.0) & 80.7 (82.0/79.4)\\ \midrule
      SciBERT & 85.6 (87.1/84.1) & 74.6 (77.2/72.3) & 81.9 (84.6/79.4) & 11.3 (19.0/8.0) & 80.3 (82.7/78.1) \\
      BioBERT & 85.1 (84.5/85.7) & 76.9 (79.3/74.6) & 82.6 (82.6/82.5) & 15.2 (16.6/14.0) & 81.0 (81.7/80.3) \\
      MatSciBERT & 85.8 (84.4/87.3) & 77.4 (78.2/76.5) & 82.4 (81.9/82.8) & 11.4 (13.2/10.0) & 81.3 (81.1/81.7)\\ \midrule
      GPT-3.5-Turbo & 63.7 (61.4/67.2) & 49.4 (47.5/52.5) & 59.5 (86.6/45.9)  &  2.2 (17.5/1.3) &  56.4 (58.8/54.1)\\
      GPT-4 & 64.7 (57.6/75.2) & 61.6 (52.2/76.6) & 74.2 (67.1/84.2) & 5.7 (8.5/4.8)& 64.5 (56.5/75.1) \\
      \bottomrule
    \end{tabular}
  \caption{
   Main NER results on the test dataset, presented as ``F-1 Score (Precision/Recall)'' in $\%$. We offer scores under different metrics for each entity category and the overall micro-average performance.
  }
  \label{tb:main.results}
\end{table*}

\subsection{Experimental Setup}

\noindent \textbf{Evaluation Protocol.}
We split the dataset into $123$ training articles, $27$ validation articles, and $27$ test articles following a $70\%/15\%/15\%$ ratio.
The three sets do not have overlapping scientific documents.
For NER, we report the entity-level precision, recall, and F-1 scores of each baseline for different entity categories, as well as the corresponding micro-average of these metrics.
For RE, we report the precision, recall, and F-1 score.

\noindent \textbf{Hyperparameters.} 
For BiLSTM-CRF, we use one layer of BiLSTM layer with $256$-dimensional hidden states and $128$ embedding dimensionality.
For the BERT-family NER models, we stack a linear layer with a hidden size of $128$ on the BERT architecture for token classification.
For all the NER and RE models, we use early stopping on the dev set for regularization. 
See App.~\ref{a1:impl} for details.


\subsection{Main Results}
\noindent \textbf{Entity Mention Extraction.}
Table~\ref{tb:main.results} shows the performance of different methods for the NER task on \ours.
From the results, we make the following observations:
(1) BERT-based models significantly outperform BiLSTM-CRF model with a $14.8\%$ gain in micro average F1-score.
This is because BERT-based models have been pre-trained on a large corpus of data, allowing them to possess more semantic knowledge than BiLSTM-CRF and to better understand the context.
(2) Domain-specific BERT models achieve slightly better performance than the vanilla BERT due to the encoding of domain-specific knowledge. 
MatSciBERT, which is fine-tuned on a corpus of materials science articles, shows the best performance on almost all metrics.
(3) Upon comparing the performance of different entity types, we find that it is challenging for all models to discriminate \texttt{Condition} entities from the other categories. 
We hypothesize that this is because \texttt{Conditions} are relatively rare in the training data, and the \texttt{Condition} entities could resemble property value entities. 

\begin{table}[h]
\centering
\small
\setlength{\tabcolsep}{0.3em}
\begin{tabular}{@{}c|c|c|c@{}}
\toprule
Model          & Precision & Recall & F-1 Score \\ \midrule
Proximity-Rule & 26.49     & 30.83  & 28.50     \\
BERT-RE        & 12.06     & 40.28  & 18.57     \\
DyGIE++        & 67.53     & 50.28  & 57.64     \\
PURE           &  60.27    &  54.04 &  56.98     \\
PURE-SUM (SciBERT)     & 42.86    & 82.50     & 56.41         \\ 
PURE-SUM (MatSciBERT)     & 51.91     & 83.06     & 63.89          \\ \midrule
GPT-3.5-Turbo & 16.37 & 34.27 & 21.73\\
GPT-4 & 37.82 & 54.16  & 44.06\\\bottomrule
\end{tabular}
\caption{
   Main RE results on the test dataset, presented as Precision, Recall, and F-1 Scores in $\%$.
  }\label{tab:re-results}
\end{table}

\noindent \textbf{Relation Extraction.}
Table~\ref{tab:re-results} shows the performance of different methods for the RE task on \ours, and we make the following observations:
(1) Among all the models evaluated, the PURE-SUM model with MatSciBERT as the encoder achieves the highest F-1 score, indicating that MatSciBERT can better understand the context, and the summation operation is an appropriate aggregation method for variable-length $N$-ary relation extraction.
(2) The rule-based approach exhibits inferior performance in comparison to most deep learning models, indicating that there are many cases that do not conform to the proximity rule, such as cross-sentence relations and parallel relations.
(3) Interestingly, the BERT-RE model shows even worse performance than the rule-based method. Compared to PURE-based models, BERT-RE directly averages the embeddings of all tokens related to the relation. As tokens with similar types have similar representations, and N-ary relations are composed of certain entity-type elements, the averaging operation results in similar relation representations, ultimately leading to poor model performance.
(4) As DyGIE++ is a model specifically designed for binary relation extraction, it can only determine the presence of N-ary relations by assessing the connectivity of arbitrary pairs of elements in the relationship.
It thus has stricter judging criteria than the other methods, making its precision higher at the cost of lower recall.

\noindent \textbf{Analysis on LLMs.}
LLMs such as GPT-3.5-turbo and GPT-4 exhibit worse performance compared to most baseline models on both NER and RE tasks. 
This discrepancy is likely due to the small proportion of polymer material science content in their pre-training corpus. 
When these models are directly prompted with few-shot examples, as opposed to being fine-tuned with training data, they receive less domain-specific information. 
This limitation hinders their ability to effectively understand and process concepts related to polymer material science.
Potential updates on LLMs, like external tools (e.g., knowledge retriever)~\cite{shi2023retrieval, zhuang2023toolqa} or collaborations between LLMs and smaller pre-trained language models~\cite{yu2023large, xu2023knowledge}, may further boost the performance via injecting more domain-specific knowledge.
Due to the poor performance obtained under the few-shot prompting setting and the high cost when fine-tuning LLMs, we recommend fine-tuning smaller domain-specific pre-trained language models, like MatSciBERT in Table~\ref{tb:main.results} and PURE-SUM (MatSciBERT) in Table~\ref{tab:re-results}, to extract polymer material science entities and relations.

\begin{figure}[!t]
	\centering
	\vspace{-1ex}
	\subfigure[Micro-Average]{
		\includegraphics[width=0.47\linewidth]{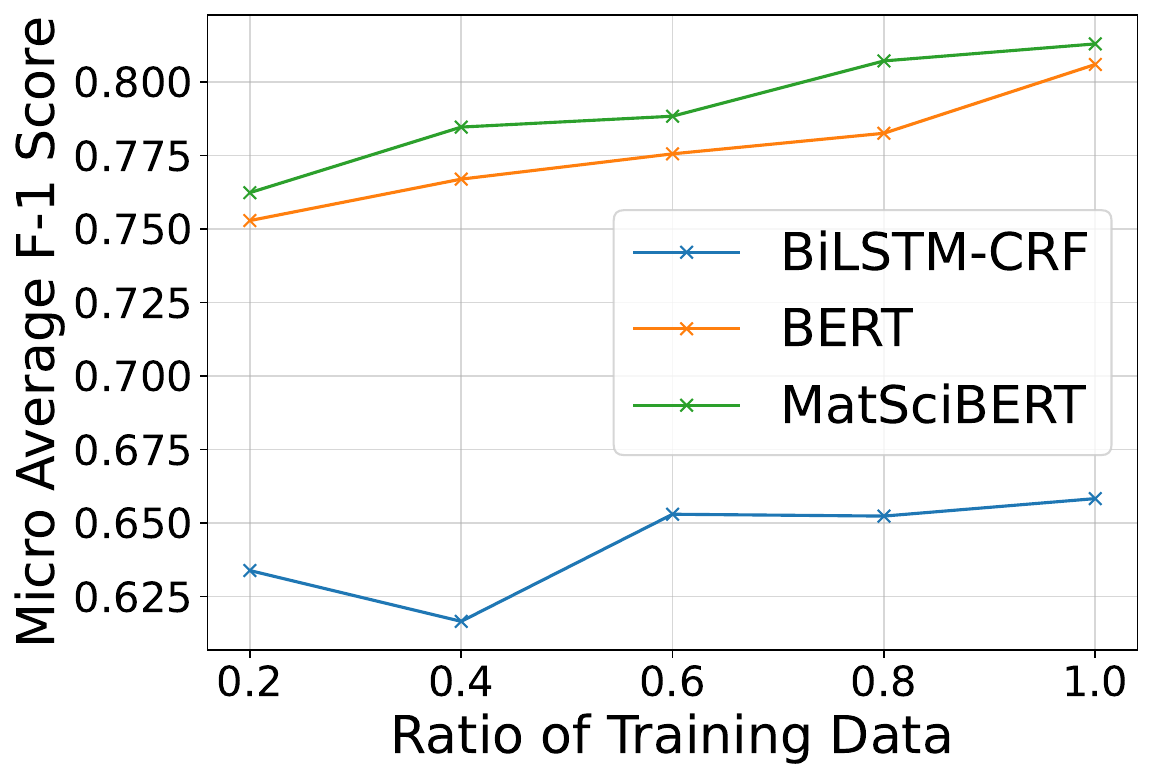}
		\label{fig:limit_f1}
	} \hspace{-2ex} 
	\subfigure[Compound Name]{
		\includegraphics[width=0.47\linewidth]{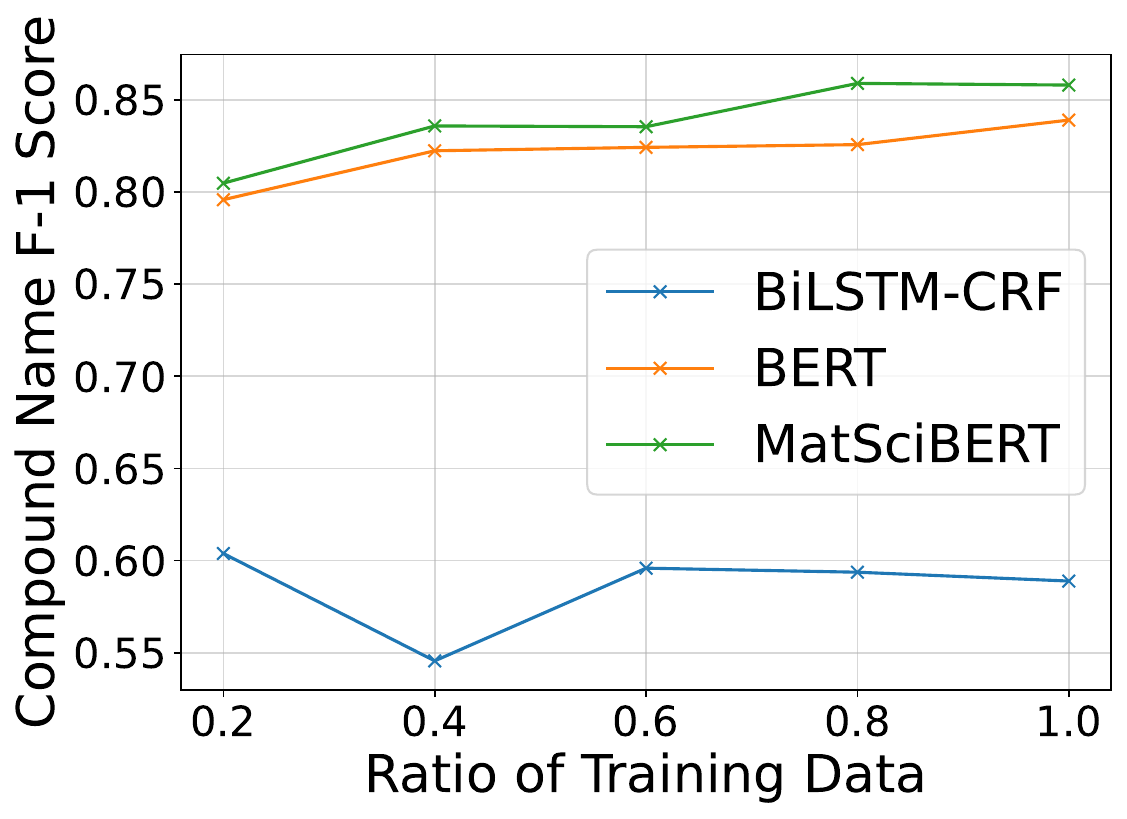}
		\label{fig:limit_cn}
	}  
 
	\subfigure[Property Name]{
		\includegraphics[width=0.47\linewidth]{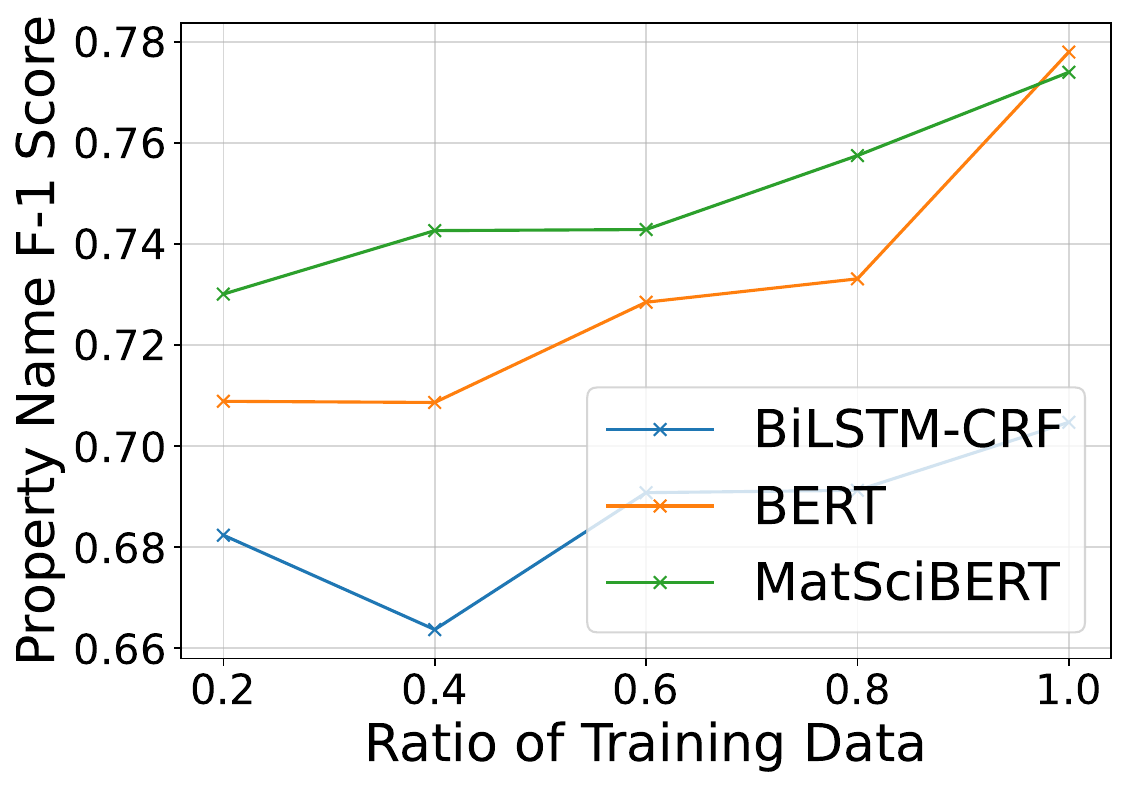}
		\label{fig:limit_pn}
	}\hspace{-1.5ex}
    \subfigure[Property Value]{
		\includegraphics[width=0.47\linewidth]{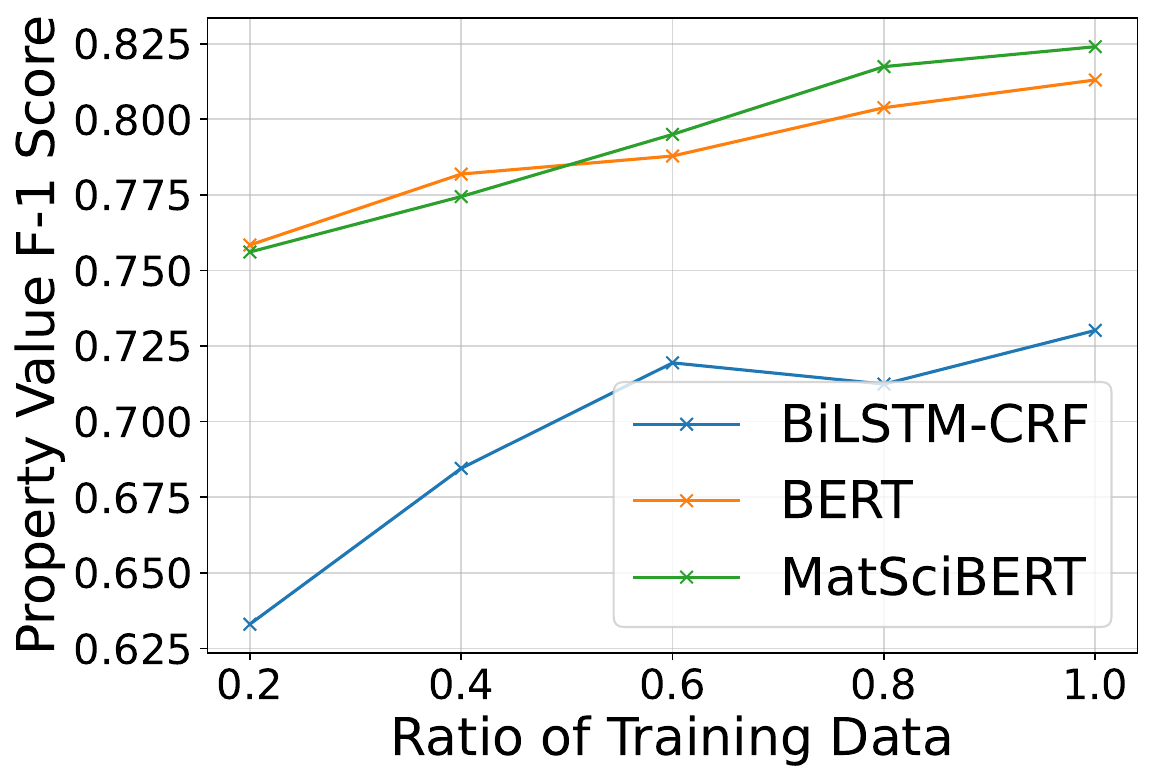}
		\label{fig:limit_pv}
	}  
	\vspace{-2ex}
	\caption{Effect of  training data size on NER task.}\label{fig:limit_data}
	\vspace{-2ex}
\end{figure}

\subsection{Impact of Data Size}

\begin{figure}[t]
	\centering
	\subfigure[Limit Data]{
		\includegraphics[width=0.47\linewidth]{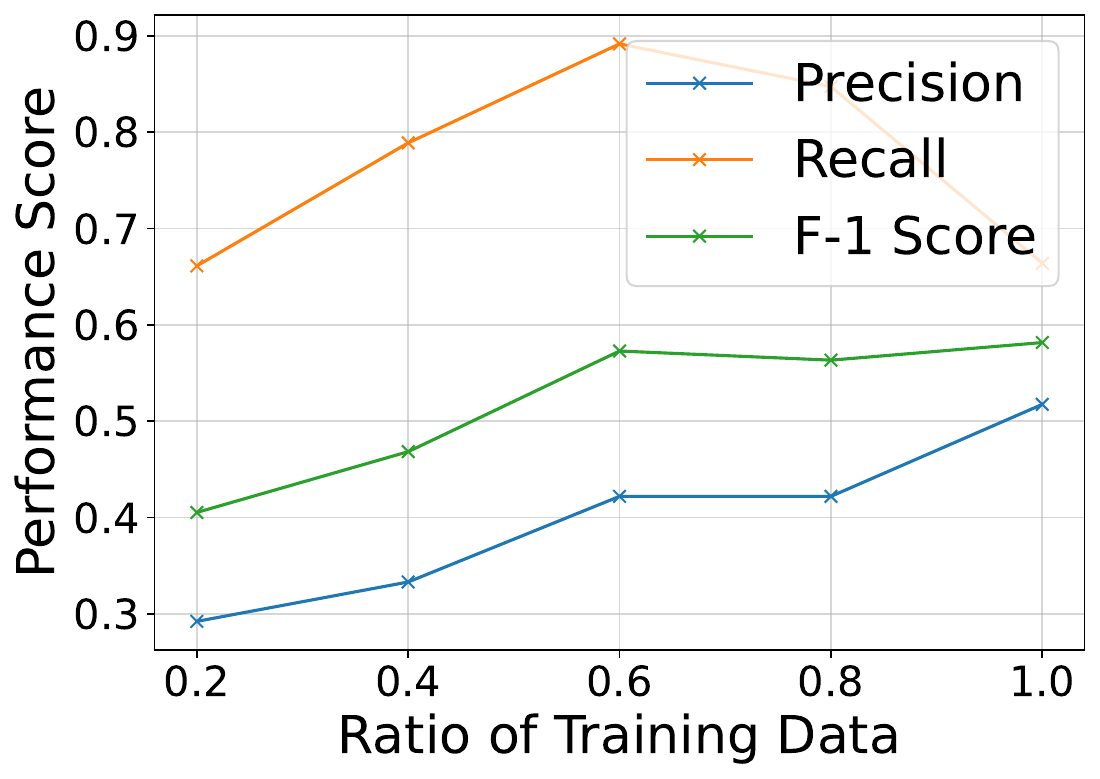}
		\label{fig:limit-re}
	} \hspace{-2ex} 
	\subfigure[Few-Shot Setting]{
		\includegraphics[width=0.47\linewidth]{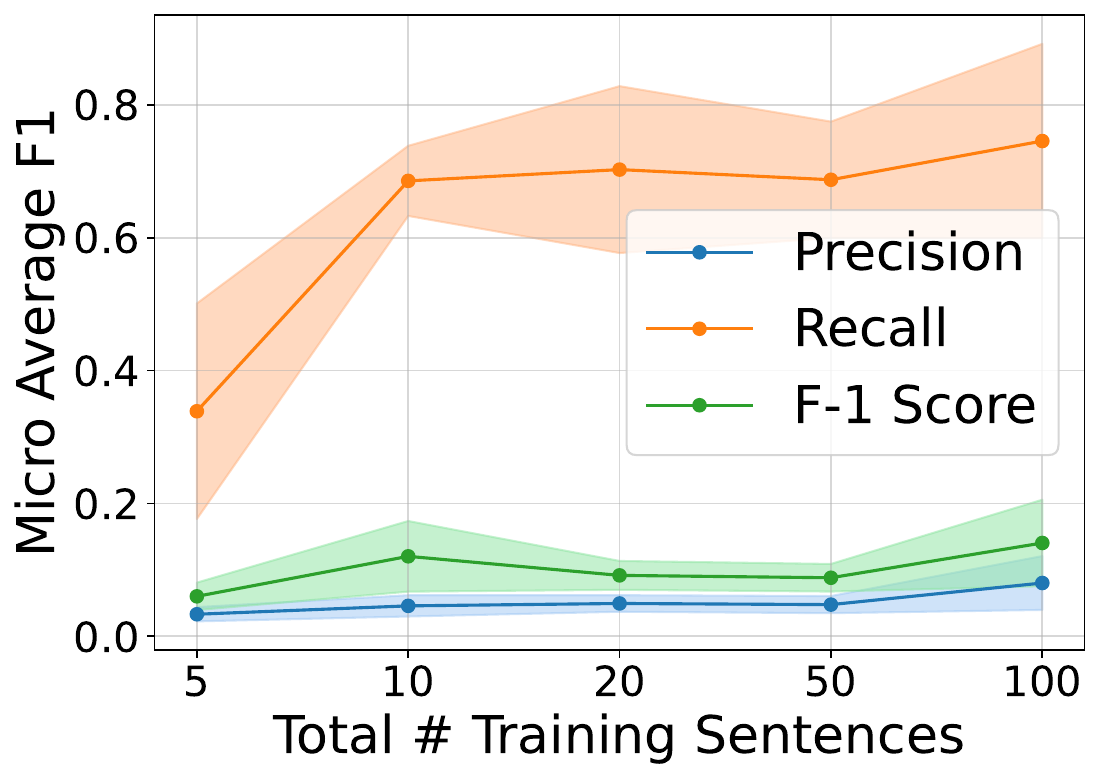}
		\label{fig:few-re}
	}  
	\vspace{-2ex}
	\caption{Performances of PURE-SUM on RE task with limited training data and few-shot setting.}\label{fig:re}
	\vspace{-2ex}
\end{figure}

\paragraph{Limited Training Data.}
We evaluate the NER model performance as a function of the amount of training data in Figure~\ref{fig:limit_data}.
Compared to BERT-based models, the performance of the BiLSTM-CRF model is consistently inferior, with only slight changes with varying sizes of training data.
This trend demonstrates the superiority of language model pre-training stage, which allows BERT-family NER models to encode relevant knowledge for the downstream task. Comparing different BERT models, MatSciBERT consistently outperforms vanilla BERT by a slight margin, which reflects the benefit of developing domain-specific pre-trained language models.


Figure~\ref{fig:limit-re} shows the performance of the best RE model PURE-SUM as training data size varies.
With more training data, the model's performance generally increases in all the metrics.
However, after training on $60\%$ data, the recall starts to decrease, while the other metrics still slightly increase.
This is because the imbalance between positive and negative cases starts to influence the training, where models are more likely to predict relations as negative, making the false negative cases increase and the recall decrease.

\paragraph{Few-Shot Learning.}
\begin{figure}[!t]
	\centering
	\vspace{-1ex}
	\subfigure[Micro-Average]{
		\includegraphics[width=0.47\linewidth]{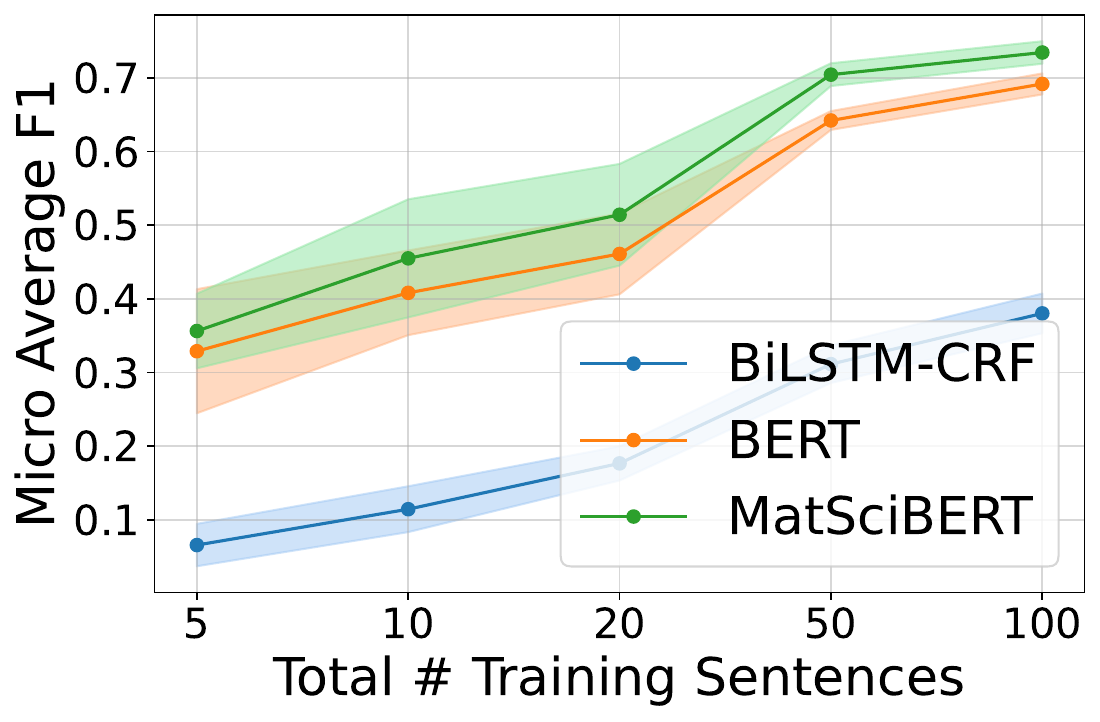}
		\label{fig:fewshot_f1}
	} \hspace{-2ex} 
	\subfigure[Compound Name]{
		\includegraphics[width=0.47\linewidth]{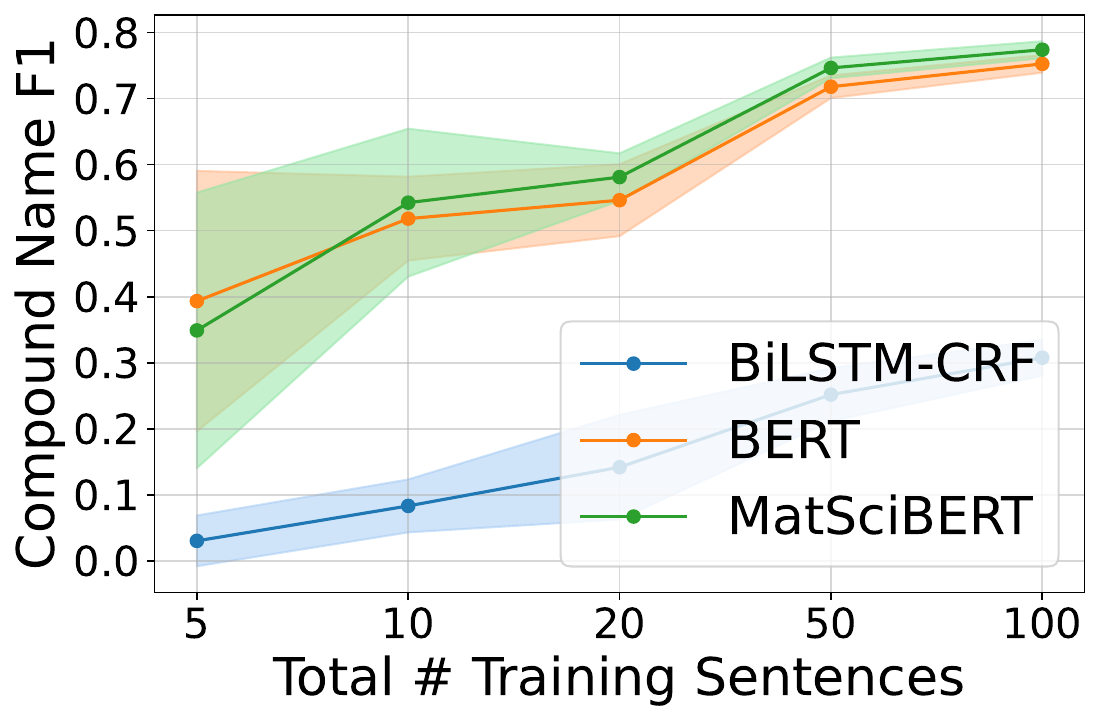}
		\label{fig:fewshot_cn}
	}  
 
	\subfigure[Property Name]{
		\includegraphics[width=0.47\linewidth]{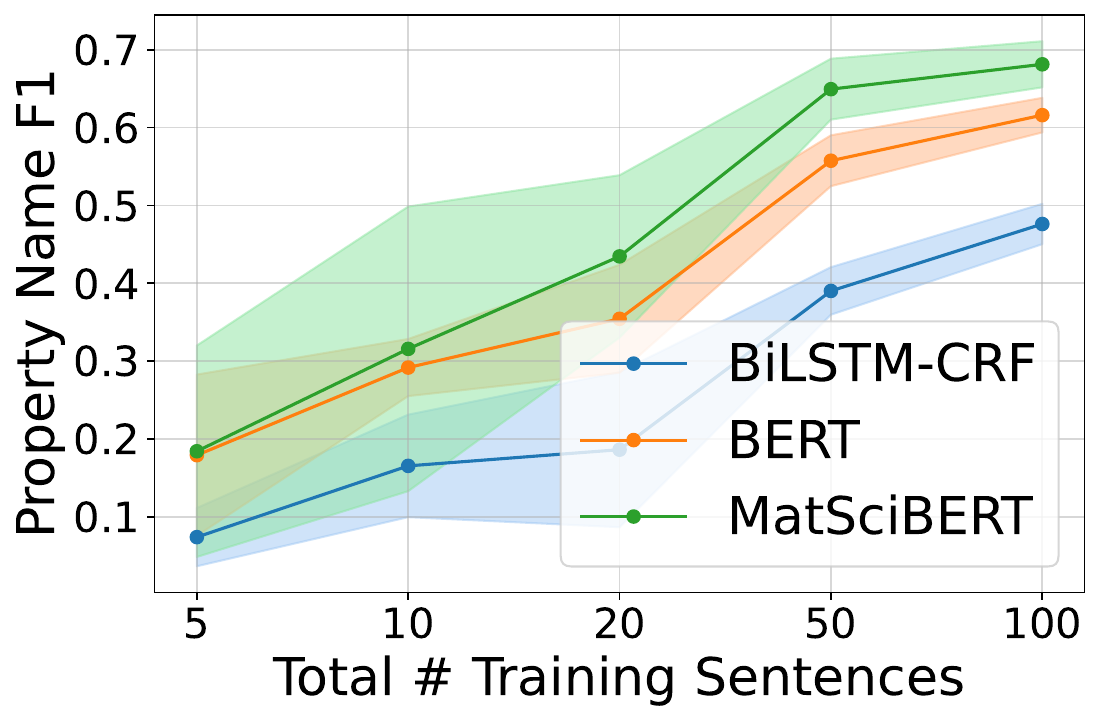}
		\label{fig:fewshot_pn}
	}\hspace{-1.5ex}
    \subfigure[Property Value]{
		\includegraphics[width=0.47\linewidth]{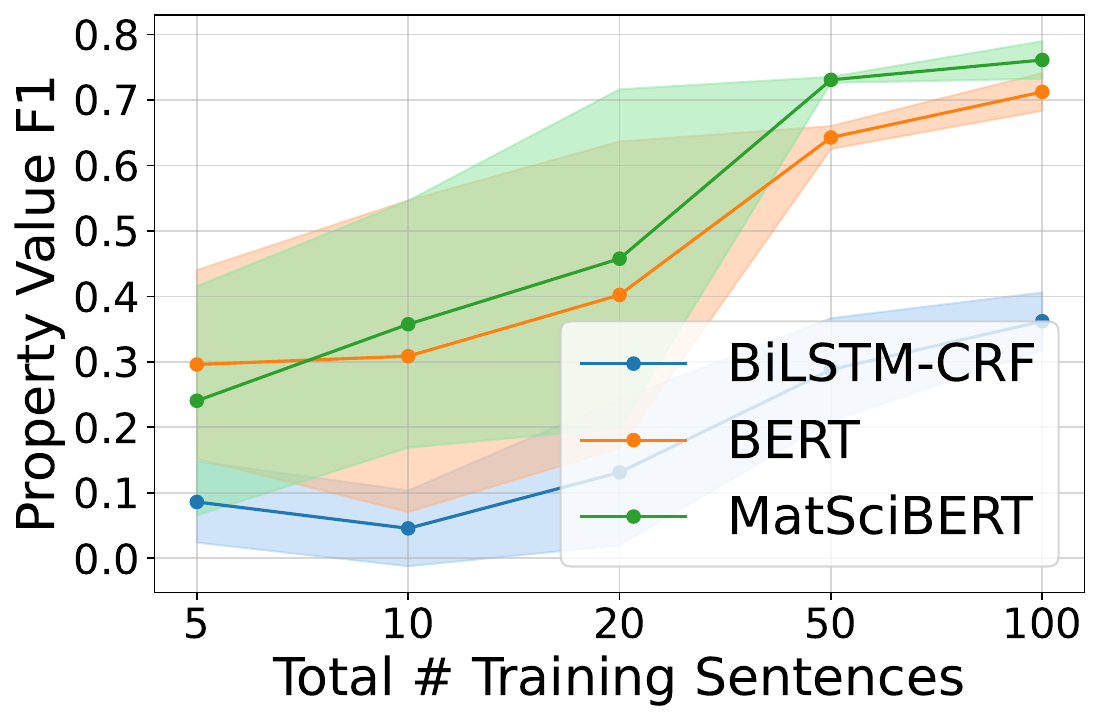}
		\label{fig:fewshot_pv}
	}  
	\vspace{-2ex}
	\caption{Effect of the few-shot learning on NER task.}\label{fig:fewshot_data}
	\vspace{-2ex}
\end{figure}

Figure~\ref{fig:fewshot_data} shows the performance of different NER models under few-shot settings. We can see BERT-based NER models consistently outperform
BiLSTM-CRF models by large margins. However, the variances of such BERT-based NER models are also much larger.
This is likely due to the different quality and representativeness of the training samples and the capacity of pre-trained language models.
The MatSciBERT model, for instance, has already captured a significant amount of domain knowledge during pre-training. When it is fed with critical cases during the fine-tuning stage, it can quickly adapt such knowledge to fine-tuning, resulting in high-quality decision boundaries on the corpus. 
However, if the training samples are of poor quality and not representative, the model's performance can be limited. 
Such instability of BERT-based fine-tuning is also observed on GLUE~\cite{mosbach2021on}.

\subsection{Error Analysis}

\begin{table*}[ht]
\centering
\footnotesize
\begin{tabular}{@{}p{0.2\linewidth}|p{0.77\linewidth}@{}} 
\toprule
Noise Types & Input Text \\ \midrule
Interweaving Relations & The corresponding HOMO and \colorbox{green!25}{\textbf{LUMO}} energy levels for \textbf{\textcolor{red}{PIDTT-TzTz}} and \colorbox{green!25}{PIDTT-TzTz-TT} are (-5.24, \colorbox{green!25}{\textbf{-3.21}}) and (-5.34, -3.03) eV, respectively.	 \\ \hline
Partially Correct Relations & For example, OFETs made using a \colorbox{green!25}{\textbf{porphyrin-diacetylene}} polymer give \colorbox{green!25}{\textbf{mobilities}} of \colorbox{green!25}{\textbf{1$\times$10$^{-7}$cm$^2$V$^{-1}$s$^{-1}$}} at \textcolor{red}{room temperature} and $2\times10^{-6}cm^2V^{-1}s^{-1}$ at $175^{\circ}$C.\\ \hline
Inverted Sentences & For polymer \colorbox{green!25}{PDTG-DPP}, the thermal stability is even better than the Sibridged analogue, PDTS-DPP, and the \colorbox{green!25}{$T_d$}=\colorbox{green!25}{409$^{\circ}$C} of PDTG-IID is the same as the Si-bridged analogue, PDTS-IID.\\
 \bottomrule
\end{tabular}
\caption{Examples incorrectly predicted by MatSciBERT. Entities highlighted in green indicate the gold $N$-ary relation in the input text. Predicted relations made by the model are shown in bold fonts. Red fonts represent the location of errors.}\label{table:re-case}
\vspace{-2mm}
\end{table*}

\begin{figure}[h]
  \centering
  \includegraphics[width=0.8\linewidth]{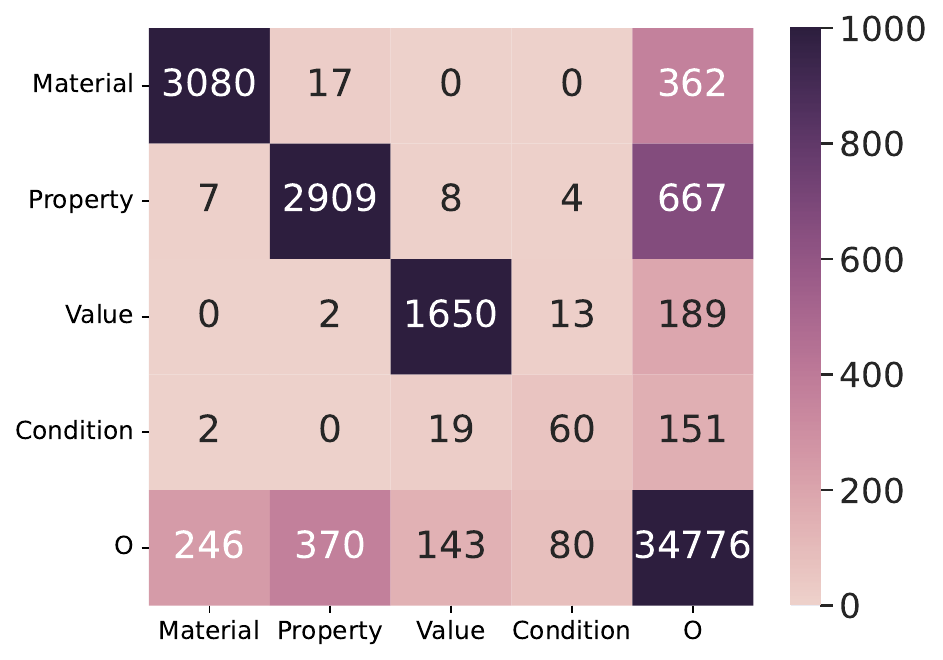}
  \caption{The confusion matrix of BERT on NER task.}
  \label{fig:confusion_matrix}
\end{figure}

We analyze the key error types of the BERT$_{\rm base}$ NER model by drawing its confusion matrix on the test set, as shown in Figure~\ref{fig:confusion_matrix}.
The confusion matrix shows that the majority of entities are correctly predicted as their gold label, with the exception of \texttt{Condition} entities.
The limited number of training samples containing \texttt{Condition} entities makes it difficult for the model to distinguish them from other irrelevant entities (labeled ``\texttt{O}'').
Additionally, the resemblance between \texttt{Condition} and \texttt{Property Value} entities often results in incorrect predictions between them.

For RE,
Table~\ref{table:re-case} illustrates the major error types made by the PURE-SUM model, including:
(1) Interweaving or parallel relations in the text present a significant challenge for models in understanding the alignment between multiple sets of entities;
(2) The task of flexible-length $N$-ary relation extraction is challenging, and errors often occur when encountering relations that cover more entities (e.g., determining whether to include the \texttt{Condition} in the prediction);
(3) The last type of error frequently arises when the sentence organization is atypical, including sentences written in the passive voice.


\subsection{Impact of Negative Sampling in Training}

\begin{figure}[h]
	\centering
	\vspace{-2ex}
	\subfigure[Quality]{
		\includegraphics[width=0.45\linewidth]{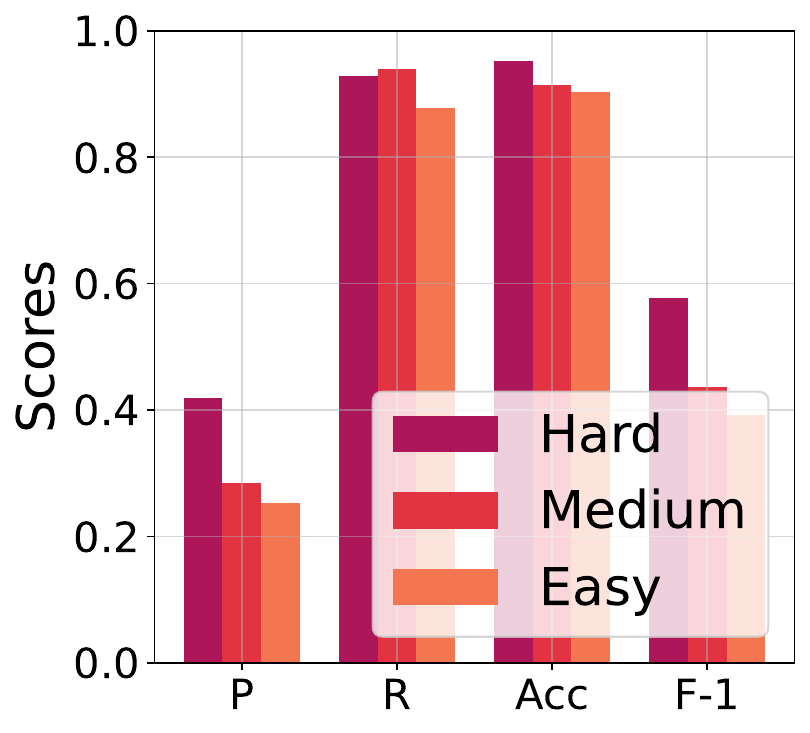}
		\label{fig:neg-qual}
	} \hspace{-2ex} 
	\subfigure[Amount]{
		\includegraphics[width=0.49\linewidth]{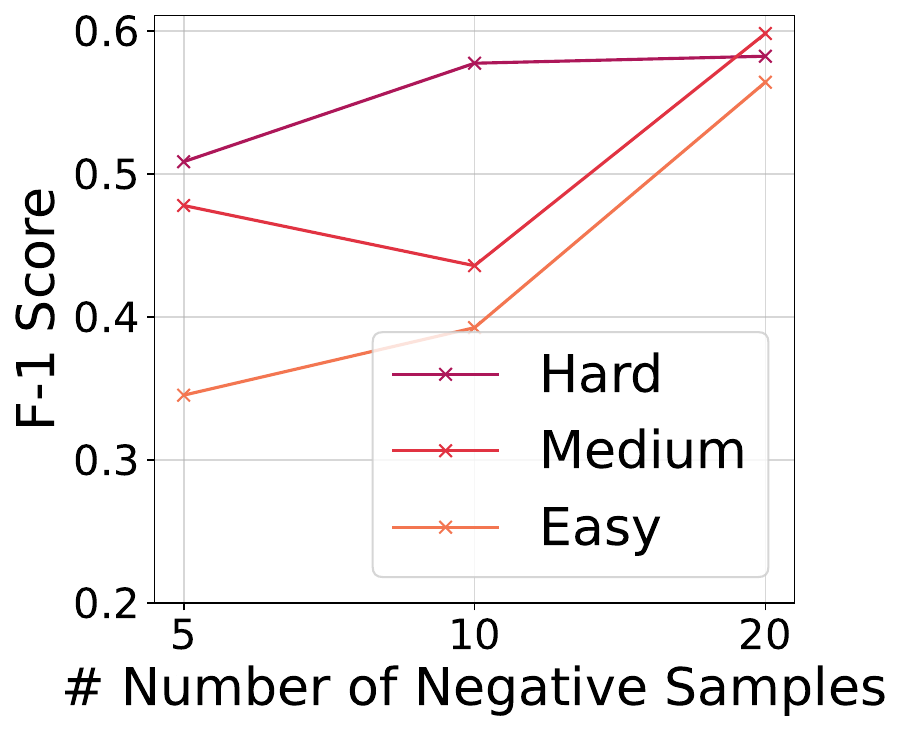}
		\label{fig:neg-amount}
	}  
	\vspace{-2ex}
	\caption{Effect of the quality and amount of negative samples during training in $N$-ary relation extraction.}\label{fig:neg}
	\vspace{-2ex}
\end{figure}

 As the RE models are trained with negative sampling, we investigate the impact of negative samples during the training process. We study three ways to create negative samples from existing relations, by corrupting entities with other irrelevant entities of the same type in the context sentences. 
(1) Easy: all possible random corruptions;
(2) Medium: single or double element corruption; 
and (3) Hard: only single-element corruption.
Figure~\ref{fig:neg-qual} shows the results when training with different negative sampling policies, with a fixed $k=10$.
We find that the hard negative sampling strategy achieves superior performance, suggesting that using hard negative cases can help the model learn better decision boundaries.
In Figure~\ref{fig:neg-amount}, we also evaluate the model performances when varying the number of negative samples $k$ from $5$ to $20$.
The trend shows that $k=20$ achieves the best performances with all different negative sampling strategies.

\section{Conclusion}
We have curated a new dataset \ours  for named entity recognition and $N$-ary relation extraction from polymer scientific literature. \ours covers thousands of 
\texttt{<Material, Property, Value, Condition>} relations curated from 146 full polymer articles.
We have evaluated  mainstay NER and RE models
on \ours and analyzed their performance and error cases. We found that even state-of-the-art models based on domain-specific pre-trained language models can struggle 
with hard NER and RE cases.
Through error analysis, we found that such difficulties arise from 
the diverse lexical formats and ambiguity of polymer named entities and also 
variable-length and cross-sentence $N$-ary relations.  Our work contributes the first 
polymer scientific information extraction dataset as well as insights into this dataset. We hope \ours will serve as a useful resource that will
and attract more research efforts from the NLP community to push the boundary of this task.


\section*{Limitations} One limitation of \ours is that we have annotated only the text modality of the polymer literature corpus.
While tables and figures are not included in \ours, they are two important modalities that contain a considerable amount of information about polymer properties.
It will be interesting to explore annotation schemes that can extend \ours to include tables and figures and enable multi-modal information extraction jointly from text, tables, and figures.
In addition, \ours currently covers four application subdomains for polymer materials.
In the future, \ours can benefit from including more sub-domains for polymers, as well as scientific publications for other organic materials.
Such extensions will not only make \ours more comprehensive for studying polymer information extraction, but also allow it to be used to study cross-domain transfer of different information extraction models.

\bibliography{custom}
\bibliographystyle{acl_natbib}

\newpage
\appendix
\section{T-SNE Visualization of Entity Embeddings}
\begin{figure*}[htb]
	\centering
	\vspace{-1ex}
	\subfigure[Original BERT$_{\rm base}$]{
		\includegraphics[width=0.32\linewidth]{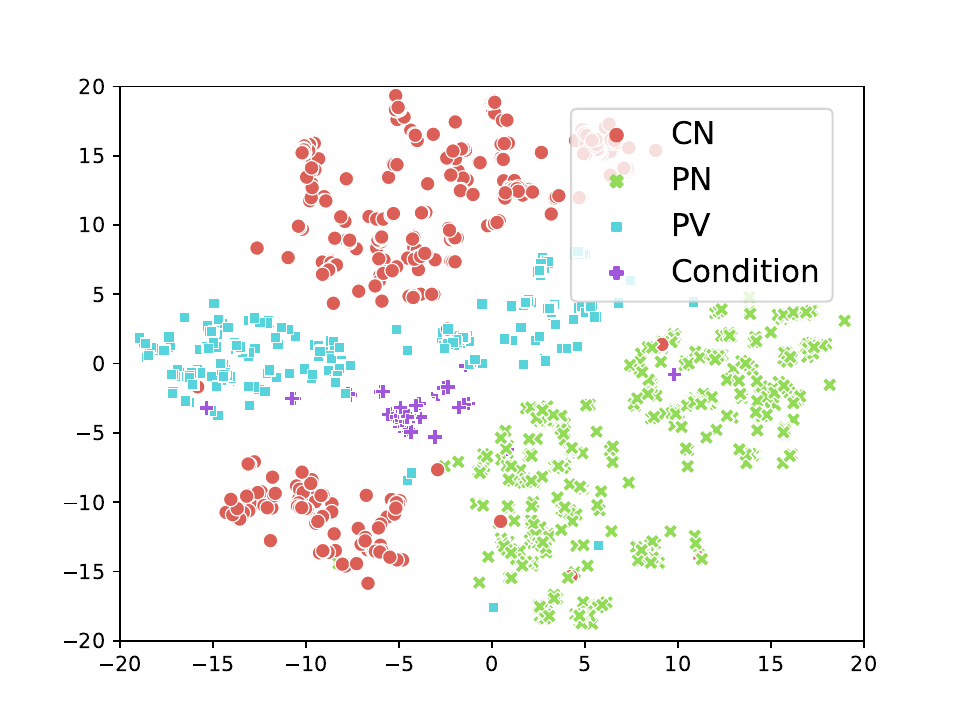}
		\label{fig:bert}
	} \hspace{-2ex} 
    \subfigure[SciBERT]{
		\includegraphics[width=0.32\linewidth]{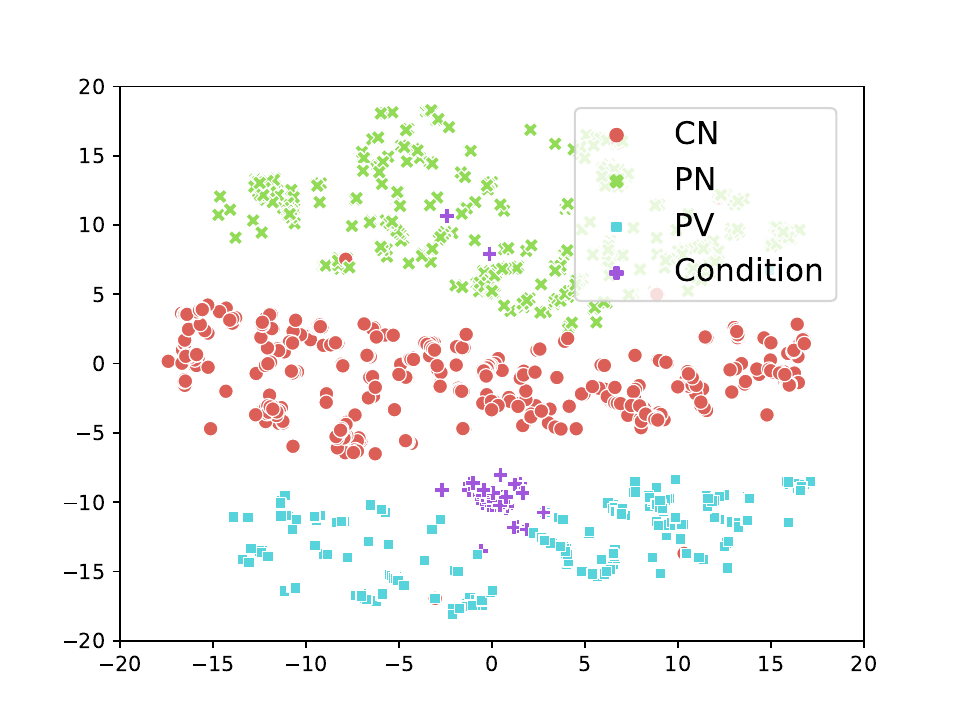}
		\label{fig:scibert}
	}  \hspace{-2ex}
	\subfigure[MatSciBERT]{
		\includegraphics[width=0.32\linewidth]{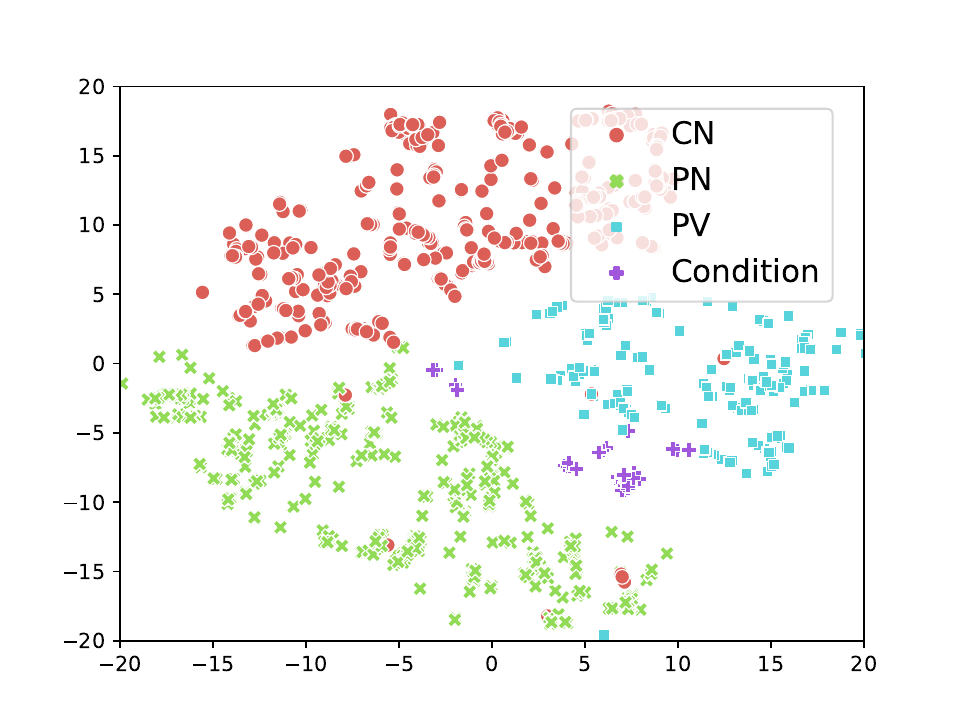}
		\label{fig:matscibert}
	} 
	\vspace{-2ex}
	\caption{t-SNE visualization of entity embeddings generated by BERT, SciBERT, and MatSciBERT.}\label{fig:entityembeds}
	\vspace{-2ex}
\end{figure*}

Figure~\ref{fig:entityembeds} shows t-SNE~\cite{van2008visualizing} visualization of entity embeddings generated by BERT$_{\rm base}$, SciBERT and MatSciBERT. 
Compared with all the visualization of different entity embeddings, we can observe that pre-training on a more similar domain of corpus to fine-tuning corpus will make model generate high-quality embeddings.
From the figures, we can easily observe that MatSciBERT embeddings of the same entity type are more clustered than those of BERT$_{\rm base}$, which is also consistent with what we observe from the quantitive results.
\section{Implementation Details}\label{a1:impl}
All the NER and RE models are trained with the Adam optimizer~\cite{kingma2014adam}, with different learning rate: 
The BiLSTM-CRF model is trained with a learning rate of $0.005$ and batch size of $64$;
While fine-tuning the BERT-family NER models, we select the learning rate of $3e-4$;
For relation extraction, instances with lengths exceeding $300$ are broken into several shorter segments, without cutting off relations, and the models are trained with a learning rate of $2e-4$ and a batch size of $8$.
All experiments are conducted on \emph{CPU}: Intel(R) Core(TM) i7-5930K CPU @ 3.50GHz and \emph{GPU}: NVIDIA GeForce RTX A5000 GPUs using python 3.8 and Pytorch 1.10. 

\section{Implementation Details of LLMs}\label{app:llm}
We conduct experiments on Azure OpenAI platform, with GPT-3.5-turbo and GPT-4 in 0613 version.
We set the temperature as $0$ to obtain a stable and faithful evaluation of the LLMs' results. 
Following the existing work~\cite{tang2023does}, we have 4 components in our NER prompt: general instruction, annotation guideline, output indicator, and few-shot exemplars. 
(1) The general instruction part specifies the objective of the LLM to mark the polymer material science entities or relations. 
(2) The annotation guideline is to provide additional explanation and guidelines for the LLM to follow when annotating different types of entities and relations. 
(3) The output indicator specifies the output format of the LLM. 
(4) The few-shot exemplars allow LLM to form a more cohesive understanding of previous instructions. 

The NER and RE prompts are presented below.

\begin{lstlisting}[caption=NER prompt.]
As a proficient linguist, your objective is to identify and label specific entities within a provided paragraph. These entities include chemical names (CN), property names (PN), property values (PV), and conditions (Condition). Chemical names, polymer material names and their abstractions are entities. Polymer material names might contain multiple chemical names within it, label them as a single entity. Abstractions of property names are also considered entities. Property values contain both the number and the unit. To represent recognized named entities in the output text, enclose them within special symbols '@', followed by their respective types '(CN)', '(PN)', or '(PV)' before the ending '@'. The remaining text should remain unchanged.
\end{lstlisting}

\begin{lstlisting}[caption=RE prompt.]
As a skilled linguist, your mission is to analyze a provided paragraph that contains four distinct types of entities: Chemical Names (CN), Property Names (PN), Property Values (PV), and Conditions (Condition). Each of these entities is enclosed within "@" symbols, with their entity type specified in brackets before the closing "@". Your objective is to identify and extract relationships among these entities, and then present them in one of two possible formats: (Chemical Names, Property Names, Property Values, Condition) or (Chemical Names, Property Names, Property Values). Please only establish relationships using the provided entities, and only provide a list of the extracted relations. Below are some examples: 
\end{lstlisting}


\section{Annotation Guidance}
In this section, we will introduce the annotation guidance. 
There are 4 types of entities that should be annotated: Chemical Compound, Property Name, Property Value, and Condition. 

\subsection{Chemical Compound} 

  \noindent $\bullet$ Only chemical nouns that can be associated with a specific structure should be labeled as Chemical Compounds: \eg,  ``4,9-di(2-octyldodecyl) aNDT'', ``trimethyltin chloride'';
  
  \noindent $\bullet$ Abbreviation of the chemical nouns should also be labeled as Chemical Compounds as long as it can be associated with a specific structure: \eg, ``PaNDTDTFBT'';
  
  \noindent $\bullet$ General chemical concepts (non-structural or non-specific chemical nouns), adjectives, verbs, and other terms that can not be associated directly with a chemical structure should not be annotated: \eg, ``polymer'', ``conjugated polymers'' should not be annotated;
  
  \noindent $\bullet$ Spans: Spans of Chemical Compounds should not contain leading or trailing spaces. If the abbreviation of Chemical Compound appears inside brackets, the brackets should not be included in the annotation. 

\subsection{Property Name}

  \noindent $\bullet$ Properties of chemical compounds should be annotated as long as they can be measured qualitatively (such as toxicity and crystallinity) or quantitatively (with a unit and a value). Property Names that occur without a corresponding value should also be annotated: \eg, ``Hole mobility'', ``Open-circuit voltage'', ``decomposition temperature'', ``conductivity'', ``toxicity'';
  
  \noindent $\bullet$ Abbreviations of Property Names should be annotated: ``PCE'', ``HOMO level'', ``LUMO level'';
  
  \noindent $\bullet$ Laboratory methods should not be annotated as Property Names: ``Titration'', ``Cyclic voltammetry'' should not be annotated as Property Names;
  
  \noindent $\bullet$ Spans: Spans of Property Names should not contain leading or trailing spaces. 

\subsection{Property Value}

\noindent $\bullet$ Both quantitative and qualitative Property Values should be annotated;

\noindent $\bullet$ Do not annotate overly vague adjectives; 

\noindent $\bullet$ Spans of Property Values should not contain leading or trailing spaces. 
Property Value and its units should be contained as a single span.
Ranges of Property Value should be contained as a single span.

\subsection{Condition}

\noindent $\bullet$ Only quantitative modifiers that constrain the numerical Property Value should be annotated as Conditions;

\noindent $\bullet$ Spans of Conditions should not contain leading or trailing spaces. 

\begin{figure*}[t]
  \centering
  \includegraphics[width=\linewidth]{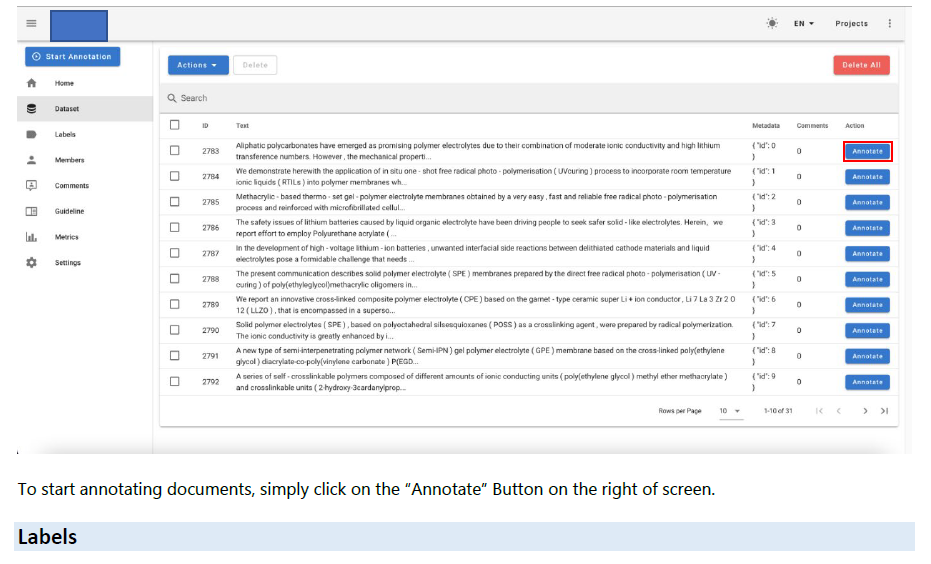}
  \caption{Overview of documents to annotate on the annotation platform.}
  \label{fig:guide3}
\end{figure*}

\begin{figure*}[t]
  \centering
  \includegraphics[width=\linewidth]{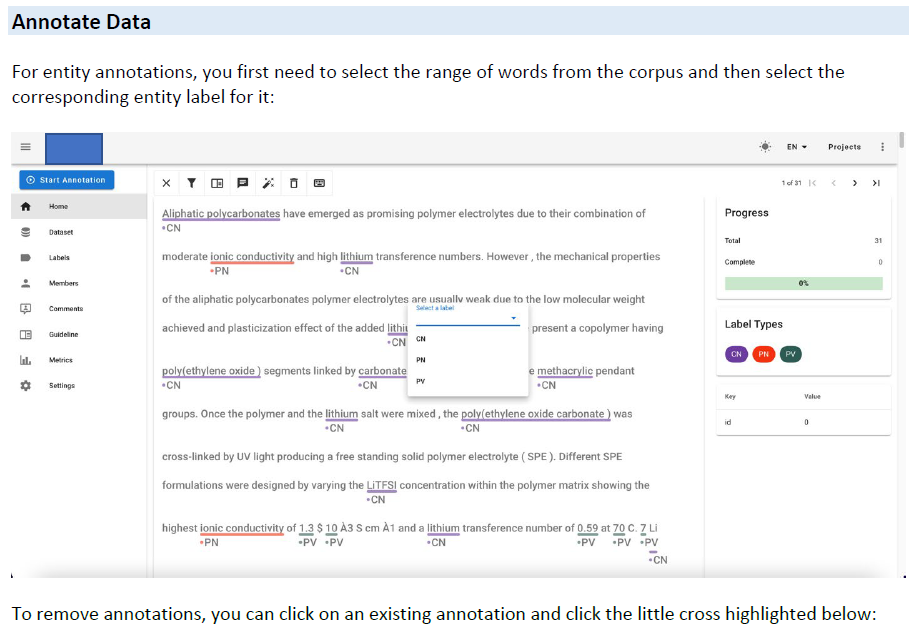}
  \caption{Instructions on assigning pre-defined labels to named entities.}
  \label{fig:guide5}
\end{figure*}

The screenshots of the official annotation guidance shared with all the annotators are listed in Figure~\ref{fig:guide3} and Figure~\ref{fig:guide5}.



\end{document}